\documentclass{article} %
\usepackage{colm2024_conference}

\usepackage{booktabs}
\usepackage{graphicx}
\usepackage{enumitem}
\usepackage{wrapfig}
\usepackage{algorithm}
\usepackage{algpseudocode}

\usepackage{microtype}
\usepackage{amsmath}
\usepackage{colortbl}
\usepackage[utf8]{inputenc}
\definecolor{lightgray}{rgb}{0.9,0.9,0.9}
\usepackage{caption}
\usepackage{subcaption}
\usepackage{xcolor}
\usepackage{setspace}
\usepackage{url}
\usepackage{multirow}
\usepackage{colortbl}
\usepackage{tabularx}
\usepackage{blindtext}
\usepackage{pgfplots}
\pgfplotsset{compat=1.18} 
\usepackage{tikz}
\usetikzlibrary{er,positioning,bayesnet}
\usepackage{makecell}
\usepackage{tipa}
\usepackage{siunitx}
\usepackage{nicefrac}
\usepackage{tocloft}
\usepackage{listings}
\usepackage[raster,skins]{tcolorbox} %
\usepackage{xltabular}
\usepackage{adjustbox}
\usepackage{xurl}

\usepackage{amsmath,amsfonts,bm}

\def\eqref#1{equation~\ref{#1}}

\def\1{\bm{1}}

\DeclareMathAlphabet{\mathsfit}{\encodingdefault}{\sfdefault}{m}{sl}
\SetMathAlphabet{\mathsfit}{bold}{\encodingdefault}{\sfdefault}{bx}{n}

\title{Ovis-U1 Technical Report}

\author{
\bf Ovis Team, Alibaba Group
}

\begin{document}

\maketitle

\begin{abstract}
In this report, we introduce Ovis-U1, a 3-billion-parameter unified model that integrates multimodal understanding, text-to-image generation, and image editing capabilities. Building on the foundation of the Ovis series, Ovis-U1 incorporates a diffusion-based visual decoder paired with a bidirectional token refiner, enabling image generation tasks comparable to leading models like GPT-4o. 
Unlike some previous models that use a frozen MLLM for generation tasks, Ovis-U1 utilizes a new unified training approach starting from a language model. 
Compared to training solely on understanding or generation tasks, unified training yields better performance, demonstrating the enhancement achieved by integrating these two tasks.
Ovis-U1 achieves a score of 69.6 on the OpenCompass Multi-modal Academic Benchmark, surpassing recent state-of-the-art models such as Ristretto-3B and SAIL-VL-1.5-2B. In text-to-image generation, it excels with scores of 83.72 and 0.89 on the DPG-Bench and GenEval benchmarks, respectively. For image editing, it achieves 4.00 and 6.42 on the ImgEdit-Bench and GEdit-Bench-EN, respectively. As the initial version of the Ovis unified model series, Ovis-U1 pushes the boundaries of multimodal understanding, generation, and editing. 
\end{abstract}

\vspace{-15pt}

\begin{figure}[hbp]
    \centering
    \includegraphics[width=0.85\textwidth]{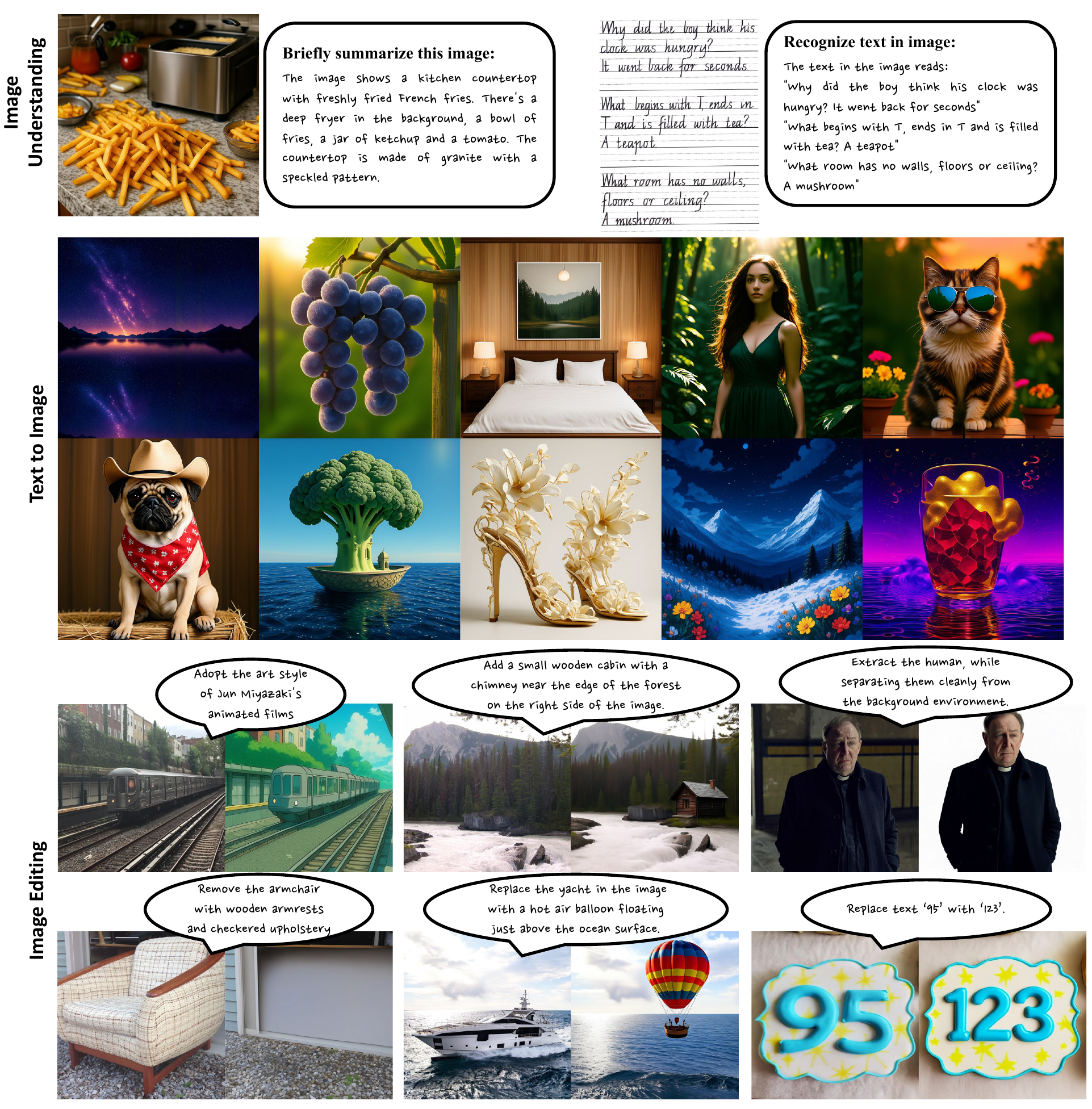}
    \caption{Comprehensive illustration of the functional capabilities of Ovis-U1.}
    \label{fig:intro}
\end{figure}

\newpage

\section{Introduction}
\label{sec:intro}

The rapid evolution of multimodal large language models (MLLMs) has been a driving force behind the increasing sophistication of artificial general intelligence (AGI). Recent developments, notably GPT-4o, introduced by \cite{chatgpt4o}, have shown that unified models capable of both understanding and generating across multiple modalities can significantly transform a wide range of real-world applications. GPT-4o integrates native image generation with advanced language capabilities, empowering users to execute complex visual tasks through natural language dialogue. These tasks (e.g., image editing \citep{brooks2023instructpix2pix}, multi-view synthesis \citep{mildenhall2021nerf}, style transfer \citep{gatys2016image}, object detection \citep{zou2023object}, instance segmentation \citep{hafiz2020survey}, depth estimation \citep{mertan2022single}, normal estimation \citep{qi2018geonet}), which previously required specialized models, can now be performed with high efficiency and accuracy. This represents a breakthrough in multimodal perception and marks the beginning of a new era where unified multimodal understanding and generation models \citep{zhang2025unified} handle both text and visual tasks seamlessly.

The emergence of GPT-4o marks a significant transition towards a unified multimodal understanding and generation framework in areas related to AGI. This raises two fundamental questions. First, how can a multimodal understanding model be endowed with the capability to generate images? This requires careful design of a visual decoder that can work seamlessly with the large multimodal language model. Second, how can a unified model be effectively trained on both understanding and generation tasks? We have observed that GPT-4o's understanding performance is enhanced by integrating image generation capabilities, suggesting that unified training may collaboratively improve performance across a range of tasks. In this report, we will study these two questions by our Ovis-U1 model.

Drawing inspiration from GPT-4o, we present Ovis-U1, a unified model with 3 billion parameters that expands the capabilities of the Ovis series \citep{lu2024ovis}. This model incorporates a novel visual decoder built on a diffusion Transformer architecture \citep{flux2024, esser2024scaling} and a bidirectional token refiner \citep{ma2024exploring, kong2024hunyuanvideo} to enhance the interaction between textual and visual embeddings. These advancements allow Ovis-U1 to generate high-quality images from textual descriptions and refine images based on textual prompts.
Ovis-U1 is trained using a unified strategy that simultaneously tackles various tasks with a diverse array of multimodal data. Comprehensive ablation studies show that our unified training approach collaboratively enhances both understanding and generation performance.

The vision for Ovis-U1 is twofold: firstly, to advance existing MLLM models by introducing novel architecture and training strategies that improve the understanding, generation, and editing of multimodal data, thereby enhancing precision and flexibility in handling complex tasks. Secondly, by open-sourcing Ovis-U1, we aim to accelerate AI development within the community, encouraging collaborative research and innovation to hasten the creation of general-purpose AI systems capable of advanced multimodal reasoning and manipulation.

In this report, the emergence of Ovis-U1 represents a significant step forward in the development of multimodal AI systems, expanding on the strengths of the Ovis series while paving the way for future advancements. Below, we show the key features of Ovis-U1:
\begin{itemize}
    \item \textbf{Diversity of Data}: Ovis-U1 has been trained on a diverse composition of multimodal data, spanning text-image understanding, text-to-image generation, and image editing tasks. This diverse training ensures that the model excels across a wide range of applications, from generating detailed images from textual descriptions to refining and editing images based on complex prompts. By learning from multiple tasks in a unified framework, Ovis-U1 achieves improved generalization, seamlessly handling real-world multimodal challenges with high accuracy.
    \item \textbf{Architecture Improvement}: Building upon the previous Ovis models, Ovis-U1 enhances its multimodal understanding capabilities by introducing a novel visual decoder based on diffusion architecture and a bidirectional token refiner to strengthen the interaction between textual and visual embeddings. The visual decoder utilizes multimodal diffusion Transformer (MMDiT) with rotary position embedding (RoPE) as the backbone, allowing for high-fidelity image generation from text. The bidirectional token refiner improves the interaction between textual and visual features, significantly enhancing text-to-image synthesis and image manipulation tasks.
    \item \textbf{Unified Training}: Unlike previous models that specialized in single tasks, Ovis-U1 adopts a unified training approach that leverages multimodal capabilities across 6 training stages, as show in \tableautorefname~\ref{tab:training}. This approach ensures that the model learns to balance and integrate knowledge across various tasks—ranging from understanding textual and visual inputs to generating and editing images. This unified framework enables Ovis-U1 to perform seamlessly across different use cases, further pushing the boundaries of multimodal AI performance.
\end{itemize}

\section{Architecture}

\begin{figure}
  \centering
  \includegraphics[width=\linewidth]{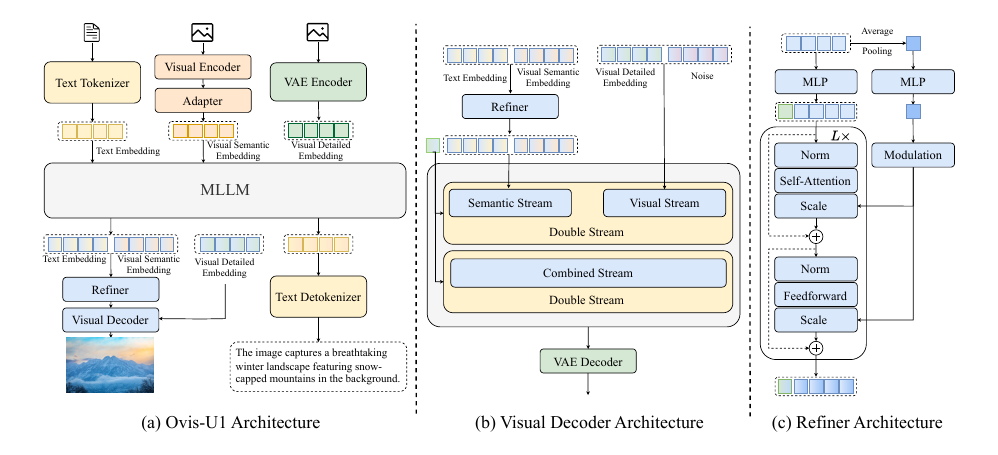}
  \caption{The overall architecture of Ovis-U1. (a) The Ovis-U1 model integrates both textual and visual inputs through a shared Multimodal Large Language Model (MLLM), using a visual decoder for image generation and a text detokenizer for text generation. An adapter bridges the vision encoder with the MLLM. A refiner module enhances the quality of the conditional embedding before decoding. (b) The architecture of the refiner module consists of two stacked Transformer blocks with modulation applied to the average pooled features. The green token represents a learnable \textit{[CLS]} token used to aggregate global information from the conditional embeddings.}
  \label{fig:framework}
\end{figure}

\begin{table}
  \caption{The model structure details of Ovis-U1.}
  \label{tab:model_detail}
  \centering
  \small
  \begin{tabular}{llll}
    \toprule
    \textbf{Module} & \textbf{\#Param. (M)} & \textbf{Pretrain} & \textbf{Trained by Stage} \\
    \midrule
    LLM & 1720 & Qwen/Qwen3-1.7B & 3 \\
    Vision Decoder & 1046 & - & 0, 4, 5 \\
    Visual Encoder & 578 & apple/aimv2-large-patch14-448 & 2, 3 \\
    Adapter & 135 & - & 1, 2, 3\\
    VAE & 84 & madebyollin/sdxl-vae-fp16-fix & \\
    Refiner & 81 & - & 0, 4, 5 \\
    \midrule
    Total & 3644 & & \\
    \bottomrule
  \end{tabular}
\end{table}

The structure of Ovis-U1 is presented by Fig.~\ref{fig:framework}. The details of each module are summarized in Tab.~\ref{tab:model_detail}. Overall, Ovis-U1 inherits the architecture of Ovis \citep{lu2024ovis} by adding a visual decoder to generate the image. 

\textbf{LLM \& Text tokenizer.} We utilize the Qwen3 series \citep{yang2025qwen3} as the backbone for the large language model. To create a unified model with 3 billion parameters, we employ Qwen3-1.7B. Unlike previous approaches that directly use a multimodal large language model (such as Qwen-VL~\citep{bai2025qwen2.5vl}) as the backbone and keep it unchanged during training, our Oivs-U1 is initialized with a language model and trained using both visual understanding and generation data. This unified training approach enhances the model's performance in both understanding and generation tasks collaboratively.

\textbf{Visual Encoder \& Adapter.}  
We enhance the visual encoder from Ovis and adopt its original visual adapter. The visual encoder, initialized from Aimv2-large-patch14-448~\citep{fini2024multimodal}, is modified to natively handle images of arbitrary resolutions, avoiding the sub-image partitioning strategy. To achieve this, we adapt the original fixed-size positional embeddings via interpolation and incorporate 2D Rotary Positional Embeddings (RoPE)~\citep{su2024roformer} for improved spatial awareness. The architecture also employs a variable-length sequence attention mechanism~\citep{dao2022flashattention, dao2023flashattention2}, following the token packing strategy from NaViT~\citep{dehghani2023patch} to efficiently process batches of images with varying resolutions. 
Following the encoder, a visual adapter bridges the vision and language modalities using the identical probabilistic tokenization scheme from Ovis. This module uses a pixel shuffle operation for spatial compression, followed by a linear head and a softmax function to transform features into a probability distribution over a visual vocabulary. The final embedding, fed to the LLM, is a weighted average from a learnable embedding table based on this distribution.

\textbf{Visual Decoder \& VAE.} We use a diffusion transformer as the visual decoder. Specifically, insparied by FLUX \citep{flux2024}, we use MMDiT~\citep{esser2024scaling} with RoPE \citep{su2024roformer} as the backbone and flow matching as the training target. By decreasing the number of layers and attention heads from 57 and 24 to 27 and 16, respectively, a 1B visual decoder is obtained. This decoder is initialized randomly and trained from scratch. Due to the limited capacity of the decoder, we employ the VAE model from SDXL with 4 channels and freeze it during the unified training. In line with FLUX.1 Redux~\citep{fluxtool2024}, the visual semantic embedding is concatenated with the text embedding to serve as semantic conditions for image generation. Additionally, following FLUX.1 Kontext~\citep{fluxkontext2024}, the context image is encoded into latent tokens using the VAE encoder. Compared to the visual semantic embedding, these context image tokens contain detailed information from the context image. Finally, these visually detailed embeddings, along with the image tokens (Noise), are input into the decoder's visual stream.

\textbf{Refiner.} We introduce a bidirectional token refiner to promote the interaction between visual embedding and textual embedding. Following \cite{kong2024hunyuanvideo, ma2024exploring}, we stack 2 transformer blocks with the modulation mechanism to consist of our refiner. Since different layers of LLM capture different levels of information about images and texts, in order to make full use of the differences in information granularity at different layers, we propose to concatenate the features of the last layer with the features of the second-to-last layer and then send them to the refiner for information interaction, which helps to generate better conditional guidance. It is worth noting that the previous text-based generation model FLUX \citep{flux2024} usually introduces CLIP to capture global features. In order to replace CLIP \citep{radford2021learning}, we introduce learnable \textit{[CLS]} token. By concatenating the learnable \textit{[CLS]} token and the embedding generated by LLM, and then sending them to the refiner for interaction, global information can be captured.

\section{Data Composition and Training Procedure}

\subsection{Data Composition}
To train Ovis-U1, we leverage three distinct types of multimodal data: multimodal understanding data, text-to-image generation data, and image+text-to-image generation data. Below, we elaborate on each category.

\noindent \textbf{Multimodal Understanding Data}.
This dataset consists of both publicly available and in-house developed data. The public datasets we utilize include COYO~\citep{kakaobrain2022coyo-700m}, Wukong~\citep{gu2022wukong}, Laion~\citep{schuhmann2022laion}, ShareGPT4V~\citep{chen2024sharegpt4v}, and CC3M~\citep{sharma2018conceptual}. Additionally, we have established a data preprocessing pipeline to filter out noisy data, enhances caption quality, and adjusts data ratio to ensure optimal training performance.

\noindent \textbf{Text-to-Image Generation Data}.
For our text-to-image generation tasks, we draw from the Laion5B dataset~\citep{schuhmann2022laion} and JourneyDB~\citep{sun2023journeydb}. Specifically, with Laion5B, we first select samples with an aesthetic score above 6. We then employ the Qwen model~\citep{Qwen2-VL} to generate detailed descriptions for each selected image, culminating in the creation of the Laion-aes6 dataset.

\noindent \textbf{Image+Text-to-Image Generation Data}.
This category can be further subdivided into four specific types:

\begin{itemize}

\item Image Editing Data: We utilize public datasets including OmniEdit~\citep{wei2024omniedit}, UltraEdit~\citep{zhao2024ultraedit}, and SeedEdit~\citep{ge2024seed}.

\item Reference-image-driven image generation Data: Our sources for this include Subjects200K~\citep{tan2024ominicontrol} and SynCD~\citep{kumari2025generating} for subject-driven image generation and StyleBooth~\citep{han2024stylebooth} for style-driven image generation.

\item Pixel-Level Controlled Image Generation Data: This encompasses tasks such as canny-to-image, depth-to-image, inpainting, and outpainting, drawing from MultiGen\_20M~\citep{qin2023unicontrol}.

\item In-House Data: We have also constructed additional datasets to complement publicly available resources, incorporating style-driven data, content removal, style translation, de-noise/de-blur data, colorization data, text rendering data, etc.
\end{itemize}

\begin{figure}
  \centering
  \includegraphics[width=0.8\linewidth]{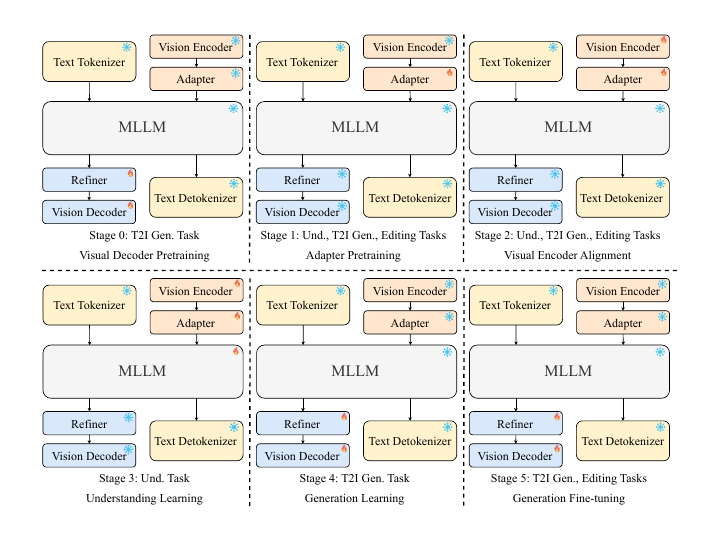}
  \caption{Overview of the proposed six-stage training pipeline. We progressively train the Ovis-U1 model through a sequence of carefully designed stages. Snowflake and flame icons denote frozen and trainable components, respectively.}
  \label{fig:training}
\end{figure}

\subsection{Training Procedure}

Different from previous works that directly using the pretrained MLLM (e.g., Qwen-VL~\citep{bai2025qwen2.5vl}), we train our model from the pretrained LLM. Given the pretrained LLM and visual encoder, Ovis has 4 training procedures totally: adapter pretraining, visual encoder alignment, understanding learning, and DPO. We add more training stages for generation. Details of each training stage are presented in Tab.~\ref{tab:training}.

\begin{table}
  \caption{Details about each training stage of Ovis-U1.}
  \label{tab:training}
  \centering
  \small
  \begin{tabular}{l|lllll}
    \toprule
    Stage & Trained Param. & Task & Steps (K) & Batch Size & Learning Rate \\
    \midrule
    0 & refiner + $Dec_i$ & T2I Gen. & 500 & 1024 & 1e-4\\
    1 & adapter & Und., T2I Gen., Editing & 1.51 & 8192 & 5e-4 \\
    2 & $Enc_i$ + adapter & Und., T2I Gen., Editing & 2.63 & 8192 & 1e-4 \\
    3 & $Enc_i$ + adapter + LLM & Und. & 23 & 2240 & 5e-5 \\
    4 & refiner + $Dec_i$ & T2I Gen. & 275 & 256 & 5e-5\\
    5 & refiner + $Dec_i$ & T2I Gen., Editing & 325 & 256 & 5e-5\\
    \bottomrule
  \end{tabular}
\end{table}

\textbf{Stage 0: Visual Decoder Pretraining.} We construct a 1B diffusion transformer for the visual decoder, starting with random initialization and training it from scratch to develop basic image generation capabilities. This stage uses text-to-image training data, enabling the visual decoder, along with the refiner, to generate images from LLM embeddings.

\textbf{Stage 1: Adapter Pretraining.} The adapter serves as a bridge between the visual encoder and LLM, aligning visual and textual embeddings. More details are provided in the Ovis paper \citep{lu2024ovis}. The adapter is randomly initialized and requires training during this stage. Unlike Ovis, Ovis-U1 is trained across understanding, text-to-image, and image editing tasks.

\textbf{Stage 2: Visual Encoder Alignment.} In this stage, both the visual encoder and adapter are fine-tuned together to further align visual and textual embeddings. Similar to Stage 1, this stage employs all three tasks for training, with the generation task aiding in the alignment of embeddings from different modalities.

\textbf{Stage 3: Understanding Learning.} This stage is the same as that of Ovis, where the parameters of the visual encoder, adapter, and LLM are trained on understanding tasks. Following this stage, these parameters are fixed to preserve the understanding capability.

\textbf{Stage 4: Generation Learning.} Since Stage 3 tunes the LLM parameters, we subsequently train the refiner and visual decoder to align with the optimized text and image embeddings. Our experiments indicate an improvement in text-to-image performance compared to Stage 0, as Stages 1-3 refine text embeddings to better align with image embeddings.

\textbf{Stage 5: Generation Fine-tuning.} Building on text-to-image capabilities, the final training stage involves fine-tuning the decoder for text-to-image and image editing tasks.

\section{Evaluation}

Like GPT-4o, recent unified multimodal models possess the ability to comprehend input images, generate images based on input prompts, and edit images according to instructions. Therefore, we benchmark the models on three tasks: image understanding, text-to-image generation, and image editing.

\textbf{Understanding.} To evaluate the understanding capabilities, we use the widely-used OpenCompass Multi-modal Academic Benchmarks\footnote{\url{https://rank.opencompass.org.cn/leaderboard-multimodal/}}, including MMBench \citep{liu2024mmbench}, MMStar \citep{chen2024mmstar}, MMMU-Val \citep{yue2024mmmu}, MathVista-Mini \citep{lu2024mathvista}, HallusionAvg \citep{guan2024hallusionbench}, AI2D-Test \citep{kembhavi2016ai2d}, OCRBench \citep{liu2024ocrbench}, MMVet \citep{yu2024mm}. The Avg Score is obtained by averaging the performance over these 8 benchmarks. Most powerful multimodal large language models have been evaluated on this Benchmark. Therefore, the unified model can compare with them conveniently.

\textbf{Text-to-Image Generation.} To evaluate the text-to-image generatation capability, we use CLIPScore \citep{hessel2021clipscore}, DPG-Bench~\citep{dpg_bench}, and GenEval~\citep{geneval} benchmarks. CLIPScore was used in DALL-E 3 \citep{betker2023improving} and the first 1K prompts\footnote{\url{https://github.com/openai/dalle3-eval-samples/blob/main/prompts/8k_coco.txt}} are intended to be used for CLIPScore calculation. DPG-Bench and GenEval are two widely-used benchmarks for text-to-image models and unified models. Some previous works rewrite prompts of GenEval to boost the performance. 
In this paper, we report the performance with the raw prompts.

\textbf{Image Editing.} To evaluate the image editing capability, we employ GEdit-Bench \citep{liu2025step1x} and ImgEdit \citep{ye2025imgedit}, two recently introduced benchmarks featuring 606 and 811 image-instruction pairs, respectively. Both benchmarks utilize the advanced GPT model to evaluate the edited images.

\begin{table}
  \caption{Performance of unified models on understanding, text-to-image generation and image editing. $\dagger$ refers to rewriting the prompt. $\times$ indicates the model is incapable of performing the task. $\ddagger$ indicates results from our own tests. }
  \label{tab:main_result}
  \small
  \centering
  \begin{tabular}{ll|lll|ll|ll}
    \toprule
    \textbf{Model} & \#Params. & \multicolumn{3}{c|}{Understanding} & \multicolumn{2}{c|}{Text-to-image Generation} &  \multicolumn{2}{c}{Image Editing} \\
     & & MMB & MMMU & MMVet & GenEval & DPG-Bench & ImgEdit & GEdit-EN \\
    \midrule
    GPT-4o & - & 86.0 & 72.9 & 76.9 & 0.84 & - & 4.2 & 7.53 \\
    \midrule
    Janus-Pro & 7B & 75.5 & 36.3 & 39.8 & 0.80 & 84.19 & $\times$ & $\times$ \\
    Emu3 & 8B & 58.5 & 31.6 & 37.2 & 0.54 & 80.60 & $\times$ & $\times$ \\
    BLIP3-o 4B & 3B + 1.4B & 78.6 & 46.6 & 60.1 & $0.81^\dagger$ & 79.36 & $\times$ & $\times$ \\
    BLIP3-o 8B &  7B + 1.4B & 83.5 & 58.6 & 66.6 & $0.84^\dagger$ & 81.60 & $\times$ & $\times$ \\
    BAGEL & 7B + 7B & 85.0 & 55.3 & 67.2 & 0.82 & 85.07 & 3.20 & 6.52 \\
    UniWorld-V1 & 7B + 12B & 83.5 & 58.6 & 67.1 & $0.84^\dagger$ & 81.38 & 3.26 & 4.85 \\
    OmniGen & 3.8B & $\times$ & $\times$ & $\times$ & 0.68 & 81.16 & 2.96 & 5.06 \\
    OmniGen2 & 3B + 4B & 79.1 & 53.1 & 61.8 & $0.86^\dagger$ & 83.57 & 3.44 & 6.42 \\
    OmniGen2$^\ddagger$ & 3B + 4B & 76.8 & 51.2 & 58.5 & - & - & - & - \\
    Ovis-U1 & 2.4B + 1.2B & 77.8 & 51.1 & 66.7 & 0.89 & 83.72& 4.00 & 6.42 \\
    \bottomrule
  \end{tabular}
\end{table}

\begin{table}
  \caption{Evaluation of understanding ability on OpenCompass Multi-modal Academic Benchmarks.}
  \label{tab:main_und}
  \centering
  \small
  \begin{tabular}{l|llllllll|l}
    \toprule
    \textbf{Model} & MMB & MMS & MMMU & MathVista & Hallusion & AI2D & OCRBench & MMVet & \textbf{Avg} \\
    \midrule
    GPT-4o & 86 & 70.2 & 72.9 & 71.6 & 57 & 86.3 & 82.2 & 76.9 & 75.4 \\
    \midrule
    InternVL2.5-2B & 70.9 & 54.3 & 43.2 & 51.1 & 42.3 & 74.9 & 80.2 & 62.6 & 59.9 \\
    SAIL-VL-2B & 73.7 & 56.5 & 44.1 & 62.8 & 45.9 & 77.4 & 83.1 & 44.2 & 61 \\
    InternVL3-2B & 78 & 61.1 & 48.7 & 57.6 & 41.9 & 78.6 & 83.1 & 67 & 61.1\\
    Qwen2.5-VL-3B & 76.8 & 56.3 & 51.2 & 61.2 & 46.6 & 81.4 & 82.8 & 60 & 64.5 \\
    Ovis2-2B & 76.9 & 56.7 & 45.6 & 64.1 & 50.2 & 82.7 & 87.3 & 58.3 & 65.2 \\
    SAIL-VL-1.5-2B & 78.5 & 62.6 & 46.4 & 67 & 50 & 83.7 & 89.1 & 58.8 & 67 \\
    Ristretto-3B & 80.2 & 62.8 & 51.3 & 67.6 & 50.2 & 84.2 & 84.7 & 60.7 & \underline{67.7} \\
    \midrule
    Ovis-U1 & 77.8 & 61.3 & 51.1 & 69.4 & 56.3 & 85.6 & 88.3 & 66.7 & \textbf{69.6} \\
    \bottomrule
  \end{tabular}
\end{table}

\begin{table}
  \caption{Evaluation of text-to-image generation ability on GenEval. $\dagger$ denotes using the rewritten prompts}
  \label{tab:main_geneval}
  \centering
  \small
  \begin{tabular}{l|cccccc|c}
    \toprule
    \textbf{Model} & Single object & Two object & Counting & Colors & Position & Attribute binding & Overall \\
    \midrule
    GPT-4o & 0.99 & 0.92 & 0.85 & 0.92 & 0.75 & 0.61 & 0.84 \\
    \midrule
    BAGEL & 0.99 & 0.94 & 0.81 & 0.88 & 0.64 & 0.63 & 0.82 \\
    $\text{BAGEL}^{\dagger}$ & 0.98 & 0.95 & 0.84 & 0.95 & 0.78 & 0.77 & 0.88 \\
    UniWorld-V1 & 0.99 & 0.93 & 0.79 & 0.89 & 0.49 & 0.70 & 0.80 \\
    $\text{UniWorld-V1}^{\dagger}$ & 0.98 & 0.93 & 0.81 & 0.89 & 0.74 & 0.71 & 0.84 \\
    OmniGen & 0.98 & 0.84 & 0.66 & 0.74 & 0.40 & 0.43 & 0.68 \\ 
    OmniGen2 & 1 & 0.95 & 0.64 & 0.88 & 0.55 & 0.76 & 0.80 \\
    $\text{OmniGen2}^{\dagger}$ & 0.99 & 0.96 & 0.74 & 0.98 & 0.71 & 0.75 & \underline{0.86} \\
    \midrule
    Ovis-U1 & 0.98 & 0.98 & 0.90 & 0.92 & 0.79 & 0.75 & \textbf{0.89} \\
    \bottomrule
  \end{tabular}
\end{table}

\begin{table}
  \caption{Evaluation of text-to-image generation ability on DPG-Bench.}
  \label{tab:main_dpg-bench}
  \centering
  \small
  \begin{tabular}{l|lllll|l}
    \toprule
    \textbf{Model} & Global & Entity & Attribute & Relation & Other & Overall \\
    \midrule
    BAGEL & 88.94 & 90.37 & 91.29 & 90.82 & 88.67 & \textbf{85.07} \\
    UniWorld-V1 & 83.64 & 88.39 & 88.44 & 89.27 & 87.22 & 81.38 \\
    OmniGen & 87.90 & 88.97 & 88.47 & 87.95 & 83.56 & 81.16 \\
    OmniGen2 & 88.81 & 88.83 & 90.18 & 89.37 & 90.27 & 83.57 \\
    \midrule
    Ovis-U1 & 82.37 & 90.08 & 88.68 & 93.35 & 85.20 & \underline{83.72}  \\
    \bottomrule
  \end{tabular}
\end{table}

\begin{table}
    \caption{Evaluation of image editing ability on ImgEdit-Bench.}
    \label{tab:main_imgedit}
    \centering
    \small
    \setlength{\tabcolsep}{4pt}
    \begin{tabular}{l|ccccccccc|c}
        \toprule
        \textbf{Model} & Add & Adjust & Extract & Replace & Remove & Background & Style & Hybrid & Action & \textbf{Overall} \\
        \midrule
        GPT-4o & 4.61 & 4.33 & 2.9 & 4.35 & 3.66 & 4.57 & 4.93 & 3.96 & 4.89 & 4.2 \\
        \midrule
        MagicBrush & 2.84 & 1.58 & 1.51 & 1.97 & 1.58 & 1.75 & 2.38 & 1.62 & 1.22 & 1.90 \\
        Instruct-Pix2Pix & 2.45 & 1.83 & 1.44 & 2.01 & 1.50 & 1.44 & 3.55 & 1.2 & 1.46 & 1.88 \\
        AnyEdit & 3.18 & 2.95 & 1.88 & 2.47 & 2.23 & 2.24 & 2.85 & 1.56 & 2.65 & 2.45 \\
        UltraEdit & 3.44 & 2.81 & 2.13 & 2.96 & 1.45 & 2.83 & 3.76 & 1.91 & 2.98 & 2.7 \\
        OmniGen & 3.47 & 3.04 & 1.71 & 2.94 & 2.43 & 3.21 & 4.19 & 2.24 & 3.38 & 2.96 \\
        Step1X-Edit & 3.88 & 3.14 & 1.76 & 3.40 & 2.41 & 3.16 & 4.63 & 2.64 & 2.52 & 3.06 \\
        ICEdit & 3.58 & 3.39 & 1.73 & 3.15 & 2.93 & 3.08 & 3.84 & 2.04 & 3.68 & 3.05 \\
        BAGEL & 3.56 & 3.31 & 1.7 & 3.3 & 2.62 & 3.24 & 4.49 & 2.38 & 4.17 & 3.2 \\
        UniWorld-V1 & 3.82 & 3.64 & 2.27 & 3.47 & 3.24 & 2.99 & 4.21 & 2.96 & 2.74 & 3.26 \\
        OmniGen2 & 3.57 & 3.06 & 1.77 & 3.74 & 3.2 & 3.57 & 4.81 & 2.52 & 4.68 & \underline{3.44} \\
        \midrule
        Ovis-U1 & 4.13 & 3.62 & 2.98 & 4.45 & 4.06 & 4.22 & 4.69 & 3.45 & 4.61 & \textbf{4.00} \\
        \bottomrule
    \end{tabular}
\end{table}

\begin{table}
    \caption{Evaluation of image editing ability on GEdit-Bench-EN.}
    \label{tab:main_gedit}
    \centering
    \tiny
    \setlength{\tabcolsep}{2pt}
    \begin{tabular}{l|ccccccccccc|c}
        \toprule
        \textbf{Model} & \makecell{Background\\Change} & \makecell{Color\\Alteration}   & \makecell{Material\\Modification}  & \makecell{Motion\\Change}  & \makecell{Portrait\\Beautification}  & \makecell{Style\\Transfer}  & \makecell{Subject\\Addition}  & \makecell{Subject\\Removal}  & \makecell{Subject\\Replacement}  & \makecell{Text\\Modification}  & \makecell{Tone\\Transformation}  & \textbf{Avg}\\
        \midrule
        GPT-4o & 7.205 &	6.491 &	6.607 & 	8.096 &	7.768 &	6.961 &	7.622 &	8.331 &	8.067 &	7.427 &	8.301 &	7.534 \\
        \midrule
        AnyEdit & 4.663	& 4.260 &	2.537 &	2.024 &	3.479	& 2.032 &	3.995 &	3.089 &	3.180 &	0.922 &	5.151 &	3.212 \\
        Instruct-Pix2Pix & 3.825 &	5.182 &	3.688 &	3.509 &	4.339 &	4.560 &	3.461 &	2.031 &	4.237 &	0.955 &	4.733 &	3.684 \\
        MagicBrush &	5.637 &	5.136 &	5.078 &	4.513 &	4.487 &	4.439 &	5.252 &	3.704 &	4.941 &	1.384 &	5.130 &	4.518 \\
        OmniGen & 5.281 &	6.003 &	5.308 &	2.916 &	3.087 &	4.903 &	6.628 &	6.352 &	5.616 &	4.519 &	5.064 &	5.062 \\
        Gemini &	6.781 &	6.369 &	6.040 &	6.938 &	5.591 &	4.676 &	7.501 &	6.447 &	7.003 &	5.765 &	6.350 &	6.315 \\
        Step1X-Edit &	6.547 &	6.545 &	6.204 &	6.483 &	6.787 &	7.221 &	6.975 &	6.512 &	7.068 &	6.921 &	6.448 &	6.701 \\
        Doubao &	7.430 &	7.095 &	6.339 &	6.973 &	6.972 &	6.767 &	7.674 &	6.748 &	7.447 &	3.471 &	7.383 &	6.754 \\
        BAGEL & 7.324 &	6.909 &	6.381 &	4.753 &	4.573 &	6.150 &	7.896 &	7.164 &	7.021 &	7.320 &	6.218 &	\underline{6.519} \\
        \midrule
        Ovis-U1 & 7.486 &	6.879 &	6.208 &	4.790 &	5.981 &	6.463 &	7.491 &	7.254 &	7.266 &	4.482 &	6.314 &	\textbf{6.420} \\
        \bottomrule
    \end{tabular}
\end{table}

\section{Experiments}

In this section, we begin by summarizing the overall performance of Ovis-U1 across understanding tasks, text-to-image generation, and image editing capabilities. Following this, we present several ablation studies to demonstrate the effectiveness of our proposed methodologies, particularly focusing on the refiner design and the performance improvements achieved through collaborative training of understanding and generation components. Finally, we showcase qualitative results to illustrate our model's capabilities.

\subsection{The main result}

Table~\ref{tab:main_result} summarizes the performance across multimodal understanding, generation, and editing tasks, comparing Ovis-U1 with models such as GPT-4o~\citep{chatgpt4o}, Janus-Pro~\citep{chen2025janus}, Emu3~\citep{wu2025less}, BLIP3-o~\citep{chen2025blip3}, BAGEL~\citep{deng2025emerging}, UniWorld-V1~\citep{lin2025uniworld}, OmniGen~\citep{xiao2024omnigen}, and OmniGen2~\citep{wu2025omnigen2}. Most of the results are sourced from OmniGen2, and we've independently tested OmniGen2’s understanding capabilities. To ensure a fair comparison, we maintain consistent generation configurations across all benchmarks. Despite having only 3.34 billion parameters, Ovis-U1 shows outstanding performance across all tasks evaluated. Detailed results for each benchmark are presented in Tables~\ref{tab:main_und} to \ref{tab:main_gedit}.

Table~\ref{tab:main_und} outlines the results on OpenCompass Multi-modal Academic Benchmarks.
Beyond methodological differences, model size significantly impacts understanding capabilities. 
We compare Ovis-U1 with the leading models within the 3B parameter range, including InternVL2.5-2B~\citep{chen2024expanding}, SAIL-VL-2B~\citep{dong2025scalable}, InternVL3-2B \citep{zhu2025internvl3}, Qwen2.5-VL-3B \citep{bai2025qwen2.5vl}, Ovis2-2B\footnote{\url{https://huggingface.co/AIDC-AI/Ovis2-2B}}, SAIL-VL-1.5-2B\footnote{\url{https://huggingface.co/BytedanceDouyinContent/SAIL-VL-1d5-2B}}, and Ristretto-3B\footnote{\url{https://huggingface.co/LiAutoAD/Ristretto-3B}}. Ovis-U1 surpasses all these models, setting a new benchmark for state-of-the-art performance, despite having only about 2B parameters dedicated to understanding tasks.

Tables~\ref{tab:main_geneval} and \ref{tab:main_dpg-bench} display the results for text-to-image generation capabilities on GenEval and DPG-Bench, respectively. We compare Ovis-U1 with recent open-source models, such as BAGEL~\citep{deng2025emerging}, UniWorld-V1~\citep{lin2025uniworld}, and OmniGen2~\citep{wu2025omnigen2}. Ovis-U1 significantly outperforms OmniGen, despite having a similar number of parameters. It's noteworthy that Ovis-U1 is equipped with a 1B visual decoder, yet it achieves performance comparable to larger models.

Tables~\ref{tab:main_imgedit} and \ref{tab:main_gedit} present the results for image editing capabilities on ImgEdit-Bench and GEdit-Bench-EN, respectively. In addition to the unified model, we also compare Ovis-U1 with various image editing models, such as MagicBrush~\citep{zhang2023magicbrush}, Instruct-Pix2Pix~\citep{brooks2023instructpix2pix}, AnyEdit~\citep{yu2025anyedit}, UltraEdit~\citep{zhao2024ultraedit}, Step1X-Edit~\citep{liu2025step1x}, and ICEdit~\citep{zhang2025context}. The performance of previous models on ImgEdit-Bench is referenced from OmniGen2~\citep{wu2025omnigen2}, while the results for GEdit-Bench-EN are sourced from the leaderboard\footnote{\url{https://step1x-edit.github.io/}}. Our model demonstrates strong performance on both benchmarks.

\subsection{Ablation study on refiner}

As shown in \tableautorefname~\ref{table:refiner}, we explore various token refiner designs for text-to-image generation tasks, comparing both clip-based and clip-free approaches. It's important to note that these ablation studies were performed on earlier versions of our model and with a limited amount of training data.
The baseline model, which combines the T5 text encoder \citep{raffel2020exploring} with the CLIP image encoder trained on about 10M text-to-image data, demonstrated solid performance with a CLIPScore of 32.19 and a DPG-Bench score of 82.32. When T5 was replaced with Qwen2.5-1.5B-Instruct \citep{an2024qwen2.5} in Variant V1, using only the last layer’s features resulted in a performance degradation, with a CLIPScore of 32.12 and a DPG-Bench score of 80.97. However, concatenating the second-to-last and last-layer features in Variant V2 restored performance to baseline levels, with scores of 32.19 and 81.48, respectively. A further enhancement was achieved by replacing Qwen2.5-1.5B-Instruct with a version fine-tuned for image-text alignment (Ovis2) in Variant V3, which led to a slight improvement in DPG-Bench (82.37) but a minor drop in CLIPScore (32.18). Moreover, the clip-free approaches were tested, with Variant V5 using a CLS token for global information outperforming Variant V4, which used averaged refiner outputs. Despite the improvements, the clip-free variants still showed slightly lower performance compared to the baseline, suggesting the potential benefits of larger datasets for better exploration of clip-free methods.

When tested on the larger 50M training data, the baseline model again outperformed other designs, with a CLIPScore of 32.57 and a DPG-Bench score of 82.97. Among the clip-free designs, Variant V7 (using CLS tokens) achieved a higher DPG-Bench score of 83.81, though its CLIPScore was slightly lower than the baseline. These findings underscore the critical role of token refiner design in LLM-based text-to-image models, highlighting how careful selection of features, particularly in the token refinement process, significantly affects the alignment of text and image information, and consequently, model performance. The results suggest that further optimizations and larger datasets are necessary to fully realize the potential of clip-free methods, especially for improving generation performance on complex benchmarks.

\begin{table}[t]
  \caption{Effect of token refiner design. Each last row refers to our final solution.}
  \label{table:refiner}
  \centering
  \small
  \begin{tabular}{l|c|ccc}
    \toprule
    \textbf{Variant} &  Data  & CLIPScore & DPG-Bench & GenEval \\
    \midrule
    Baseline (T5+CLIP) & $\sim$10M & 32.19 &	82.32  & 0.63 \\
    (V1) Last-Layer Features & $\sim$10M & 32.12 & 80.97 & 0.62 \\
    (V2) Concatenated Layer Features & $\sim$10M & 32.19 & 81.48 & 0.63 \\
    (V3) Image-Text Aligned Features & $\sim$10M & 32.18 & 82.37 & 0.61 \\
    (V4) Clip-Free - Averaged Features & $\sim$10M & 31.99 & 81.64 & 0.63 \\
    (V5) Clip-Free - CLS Token & $\sim$10M & 32.13 & 81.91 & 0.61 \\
   \midrule    
    Baseline (T5+CLIP) & $\sim$50M & 32.57 &	82.97 & 0.69 \\
    (V6) Clip-Free - Averaged Features & $\sim$50M & 32.47 & 82.65 & 0.71 \\
    (V7) Clip-Free - CLS Token & $\sim$50M & 32.42 & 83.81 & 0.67 \\
    \bottomrule
  \end{tabular}
\end{table}

\subsection{Enhance understanding with unified training}

\begin{table}
  \caption{Evaluation of understanding ability. Our unified training can enhance the understanding performance. }
  \label{tab:und}
  \centering
  \small
  \setlength{\tabcolsep}{5pt}
  \begin{tabular}{l|llllllll|l}
    \toprule
    \textbf{Variant} & MMB & MMS & MMMU & MathVista & Hallusion & AI2D & OCRBench & MMVet & \textbf{Avg} \\
    \midrule
    Baseline & 75.8 & 55.4 & 45.0 & 63.0 & 47.9 & 82.3 & 87.4 & 49.8 & 63.33 \\
    Unified Training & 77.0 & 56.9 & 44.6 & 66.3 & 46.7 & 83.7 & 87.9 & 52.8 & 64.47 \\
    \bottomrule
  \end{tabular}
\end{table}

Tab.~\ref{tab:und} presents the detailed results from the OpenCompass Multi-modal Academic Benchmark. 
We use Ovis without the unified training as our baseline for comparison. 
Compared to this baseline, Ovis-U1 demonstrates a 1.14-point improvement in average score. This enhancement validates the effectiveness of leveraging text-to-image generation and image editing tasks for aligning the visual encoder during training stages 1 and 2. It is worth noting that most previous unified models typically underperform compared to their MLLM backbones. For instance, Ming-Lite-Uni \citep{gong2025ming}, which utilizes Qwen2.5-VL-7B \citep{bai2025qwen2.5vl} as its backbone, achieves lower understanding performance. Some previous approaches~\citep{chen2025blip3,lin2025uniworld} keep the MLLM fixed, which misses the chance to enhance understanding performance.

\subsection{Enhance generation with unified training}

\begin{table}[tp]
  \caption{Performance on DPG-Bench with different training stages. The generation ability is improved progressively across different training stages. Note that the results were obtained using an early version of our model.}
  \label{tab:dpg-bench}
  \centering
  \small
  \begin{tabular}{l|lllll|l}
    \toprule
    \textbf{Model} & Global & Entity & Attribute & Relation & Other & \textbf{Overall} \\
    \midrule
    Stage1 & 80.85 & 89.56 & 89.22 & 93.54 & 80.00 & 83.81  \\
    Stage4 & 84.50 & 90.40 & 89.92 & 94.12 & 78.40 & 84.66  \\
    Stage5 & 83.59 & 90.81 & 89.46 & 94.31 & 80.80 & 85.43  \\
    \bottomrule
  \end{tabular}
\end{table}

\begin{table}[tp]
  \caption{Performance on GenEval with different training stages. The generation ability is improved progressively across different training stages. Note that the results were obtained using an early version of our model.}
  \label{tab:geneval}
  \centering
  \small
  \begin{tabular}{l|cccccc|c}
    \toprule
    \textbf{Model} & Single object & Two object & Counting & Colors & Position & Attribute binding & \textbf{Overall} \\
    \midrule
    Stage1 & 0.99 & 0.68 & 0.46 & 0.83 & 0.58 & 0.51 & 0.67  \\
    Stage4 & 0.99 & 0.77 & 0.48 & 0.86 & 0.55 & 0.58 & 0.70  \\
    Stage5 & 0.98 & 0.75 & 0.54 & 0.85 & 0.44 & 0.60 & 0.69  \\
    \bottomrule
  \end{tabular}
\end{table}

Tab.~\ref{tab:dpg-bench} and \ref{tab:geneval} summaries the image generation performance at various training stages. It's important to note that these ablation studies were performed on earlier versions of our model. In Stage 1, the model is trained using comprehensive text-to-image data, resulting in impressive performance outcomes. Following the alignment of visual and textual embeddings, Stage 4 further enhances the model's generation capabilities. In Stage 5, both text-to-image and image editing data are utilized. Notably, the inclusion of image editing data leads to a 0.77 improvement in the text-to-image performance on the DPG-Bench.

\subsection{Instruction based image editing}

\begin{figure}
  \centering
  \includegraphics[width=\linewidth]{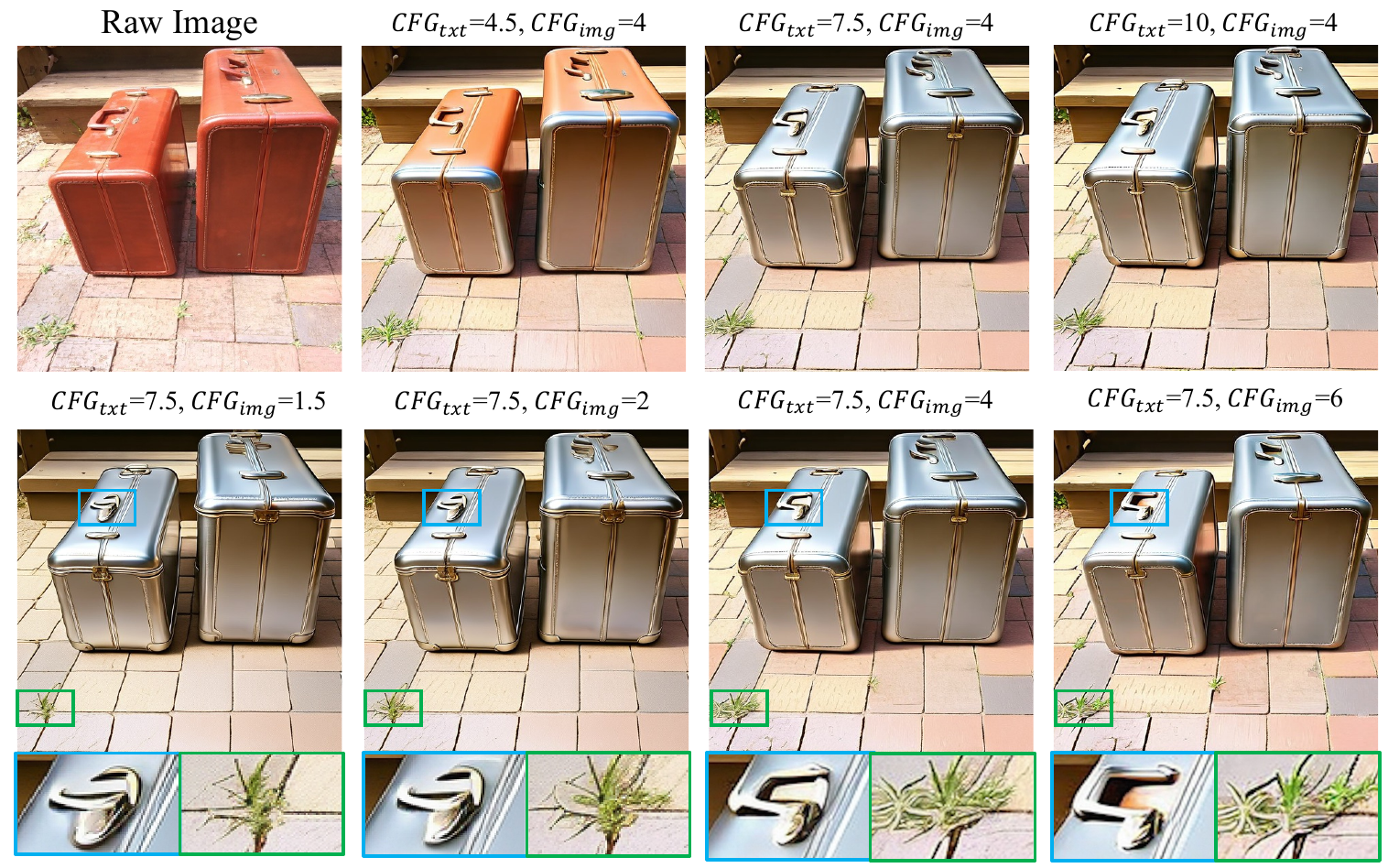}
  \caption{Qualitative results of different classifier free guidance on image editing. The instruction of editing is \textit{``change the color of bags to silver''}.}
  \label{fig:edit_cfg}
\end{figure}

\begin{table}[tp]
    \caption{Performance on ImgEdit-Bench with different classifier free guidance. Note that the results were obtained using an early version of our model.}
    \label{tab:imgedit_cfg}
    \centering
    \setlength{\tabcolsep}{4pt}
    \small
    \begin{tabular}{ll|ccccccccc|c}
        \toprule
        $CFG_{img}$ & $CFG_{txt}$ & Add & Adjust & Extract & Replace & Remove & Background & Style & Hybrid & Action & \textbf{Overall} \\
        \midrule
        1.5 & 7.5 & 4.13 & 3.57 & 3.18 & 4.49 & 4.14 & 4.30 & 4.71 & 3.72 & 4.61 & 4.06 \\
        2 & 7.5 & 4.19 & 3.76 & 3.32 & 4.41 & 4.26 & 4.36 & 4.67 & 3.84 & 4.63 & 4.13 \\
        4 & 7.5 & 4.19 & 3.66 & 3.34 & 4.40 & 4.13 & 4.25 & 4.73 & 3.55 & 4.64 & 4.09 \\
        6 & 7.5 & 4.11 & 3.67 & 3.07 & 4.29 & 3.86 & 4.24 & 4.74 & 3.30 & 4.52 & 3.98 \\
        \midrule
        4 & 5 & 4.19 & 3.68 & 3.32 & 4.31 & 3.84 & 4.28 & 4.78 & 3.45 & 4.40 & 4.04 \\
        4 & 7.5 & 4.19 & 3.66 & 3.34 & 4.40 & 4.13 & 4.25 & 4.73 & 3.55 & 4.64 & 4.09  \\
        4 & 10 & 4.02 & 3.66 & 3.31 & 4.27 & 4.13 & 4.24 & 4.74 & 3.77 & 4.54 & 4.05 \\
        \bottomrule
    \end{tabular}
\end{table}

\begin{table}[tbp]
    \caption{Performance on GEdit-Bench-EN with different classifier free guidance. Note that the results were obtained using an early version of our model. }
    \label{tab:gedit_cfg}
    \centering
    \tiny
    \setlength{\tabcolsep}{2pt}
    \begin{tabular}{ll|ccccccccccc|c}
        \toprule
        $CFG_{img}$ & $CFG_{txt}$ & \makecell{Background\\Change} & \makecell{Color\\Alteration}   & \makecell{Material\\Modification}  & \makecell{Motion\\Change}  & \makecell{Portrait\\Beautification}  & \makecell{Style\\Transfer}  & \makecell{Subject\\Addition}  & \makecell{Subject\\Removal}  & \makecell{Subject\\Replacement}  & \makecell{Text\\Modification}  & \makecell{Tone\\Transformation}  & \textbf{Avg}\\
        \midrule
        1.5 & 7.5 & 7.538 &	6.265 &	6.095 &	4.607 &	5.909 &	6.660 &	6.702 &	6.902 &	7.029 &	4.120 &	5.761 &	6.144 \\
        2 & 7.5 & 7.674 &	6.544 &	6.064 &	4.536 &	5.696 &	6.394 &	6.668 &	6.945 &	7.203 &	4.320 &	6.208 &	6.205 \\
        4 & 7.5 & 7.706 &	7.056 &	6.177 &	4.248 &	5.770 &	6.595 &	7.143 &	7.209 &	7.002 &	4.449 &	6.499 &	6.351 \\
        6 & 7.5 & 7.573 &	6.829 &	6.025 &	3.915 &	5.636 &	6.508 &	7.166 &	6.857 &	7.344 &	4.327 &	6.620 &	6.254 \\
        \midrule
        4 & 5 & 7.768 &	7.020 &	6.477 &	3.874 &	5.772 &	6.697 &	7.116 &	7.236 &	7.105 &	4.254 &	6.823 &	6.377 \\
        4 & 7.5 & 7.706 &	7.056 &	6.177 &	4.248 &	5.770 &	6.595 &	7.143 &	7.209 &	7.002 &	4.449 &	6.499 &	6.351 \\
        4 & 10 & 7.654 &	6.762 &	6.074 &	4.654 &	6.083 &	6.535 &	7.252 &	7.248 &	7.185 &	4.222 &	6.216 &	6.353 \\
        \bottomrule
    \end{tabular}
\end{table}

Following the approach of InstructPix2Pix \citep{brooks2023instructpix2pix}, we implement classifier-free guidance for both text and image conditions. Fig.~\ref{fig:edit_cfg} illustrates the results of image editing under various classifier-free guidance settings. A higher $CFG_{img}$ value preserves more details from the input image in the generated output (see blue and green boxes in Fig.~\ref{fig:edit_cfg} for details), while a higher $CFG_{txt}$ value enhances the model's adherence to the editing instructions. Quantitative evaluation results are summarized in Table~\ref{tab:imgedit_cfg} and \ref{tab:gedit_cfg}. Overall, our model demonstrates robustness to variations in the value of CFG, with result differences remaining within 0.2 for both ImgEdit-Bench and GEdit-Bench-EN. Additionally, the optimal CFG settings vary across different benchmarks. It's worth noting that our model achieves a score of 4.13 on ImgEdit-Bench with $CFG_{img}$ set to 2 and $CFG_{txt}$ set to 7.5, which is higher than the score reported in Tab.~\ref{tab:main_result}. This discrepancy is because, in Tab.~\ref{tab:main_result}, the same CFG settings are applied across all benchmarks.

\begin{figure}[tbp]
    \centering
    \includegraphics[width=0.99\textwidth]{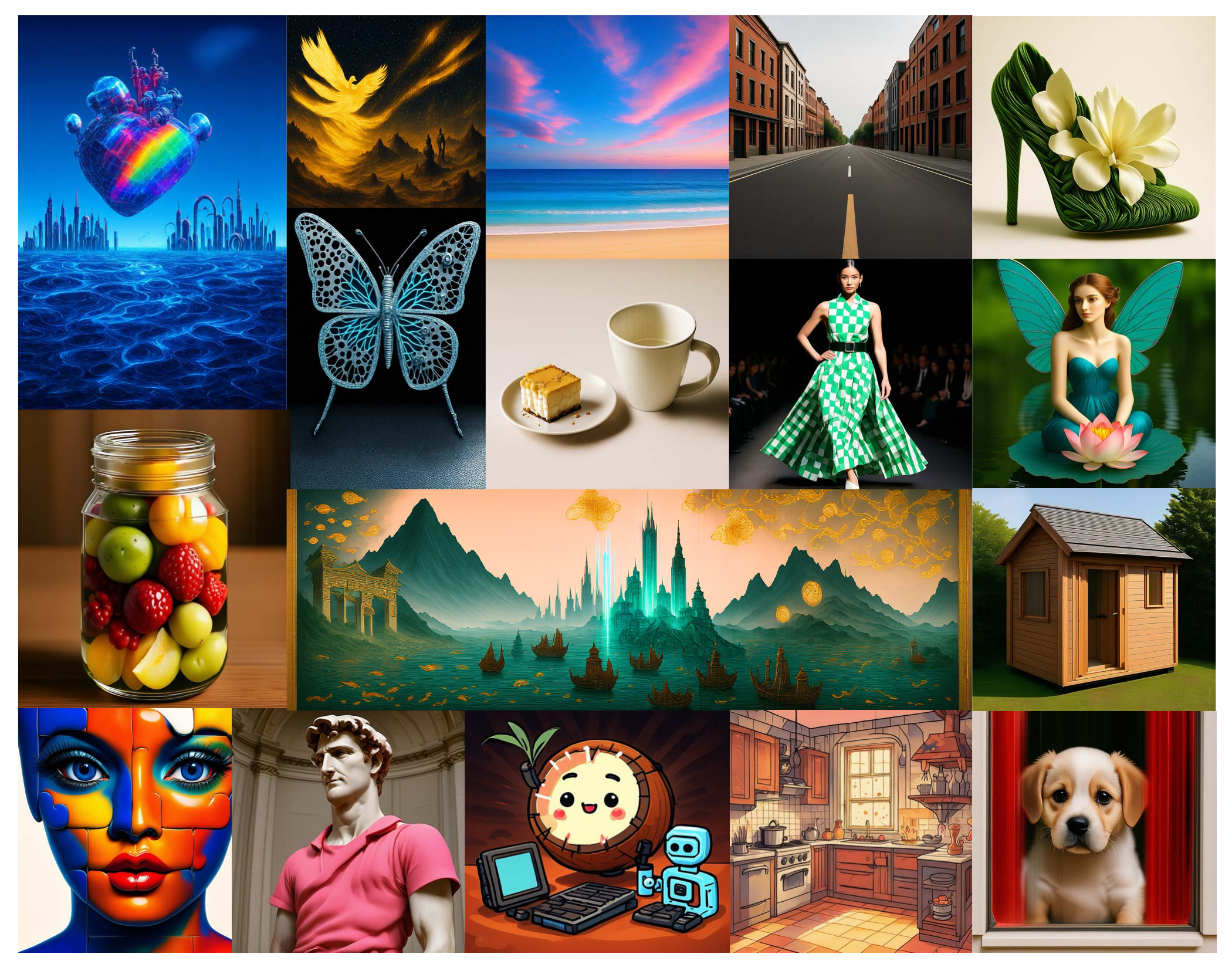}
    \caption{More qualitative results from Ovis-U1 on text-to-image generation.}
    \label{fig:qualitative_t2i}
\end{figure}

\begin{figure}[tbp]
    \centering
    \includegraphics[width=0.99\textwidth]{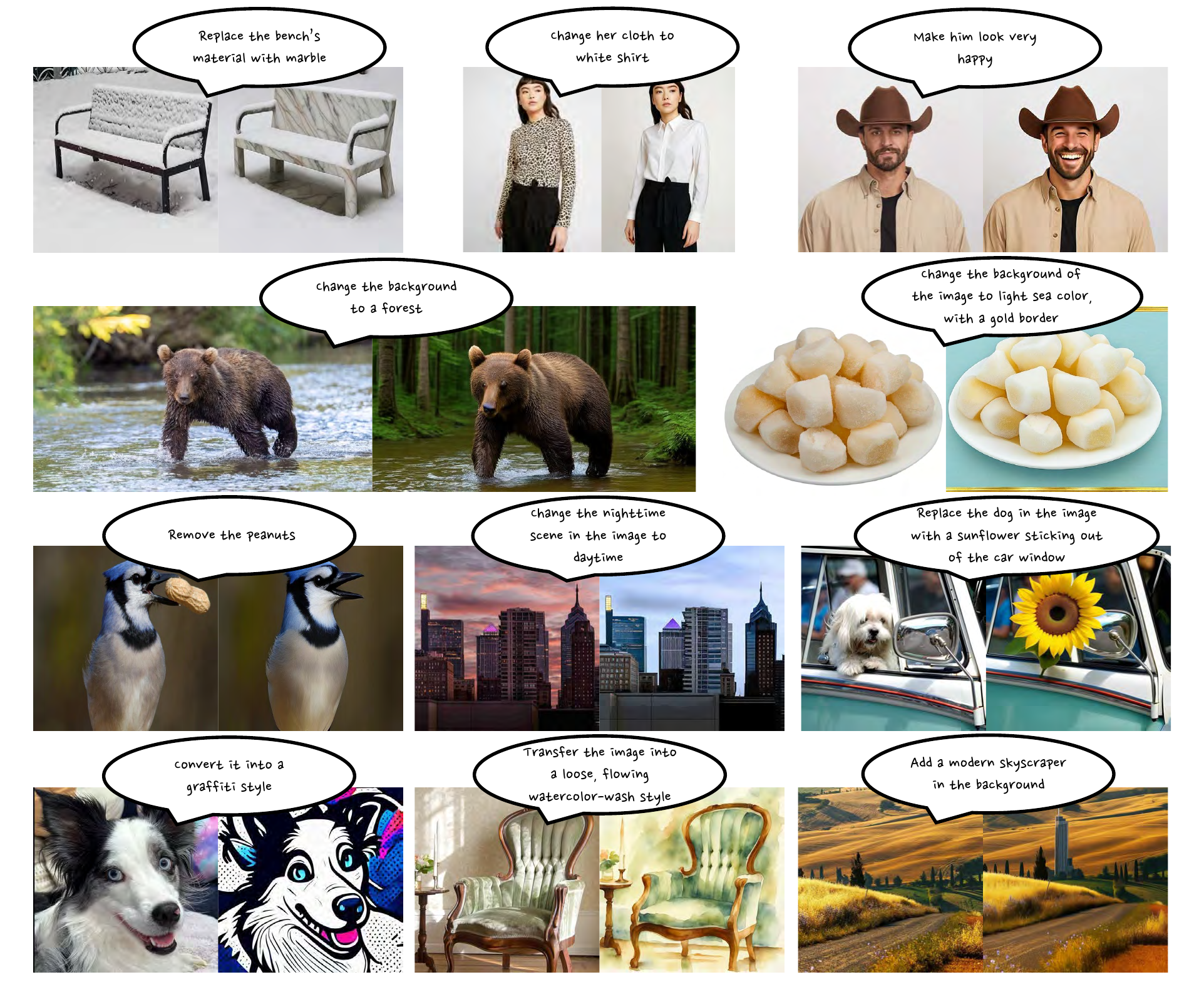}
    \caption{More qualitative results from Ovis-U1 on image editing.}
    \label{fig:qualitative_ie}
\end{figure}

\begin{figure}[tbp]
    \centering
    \includegraphics[width=0.95\textwidth]{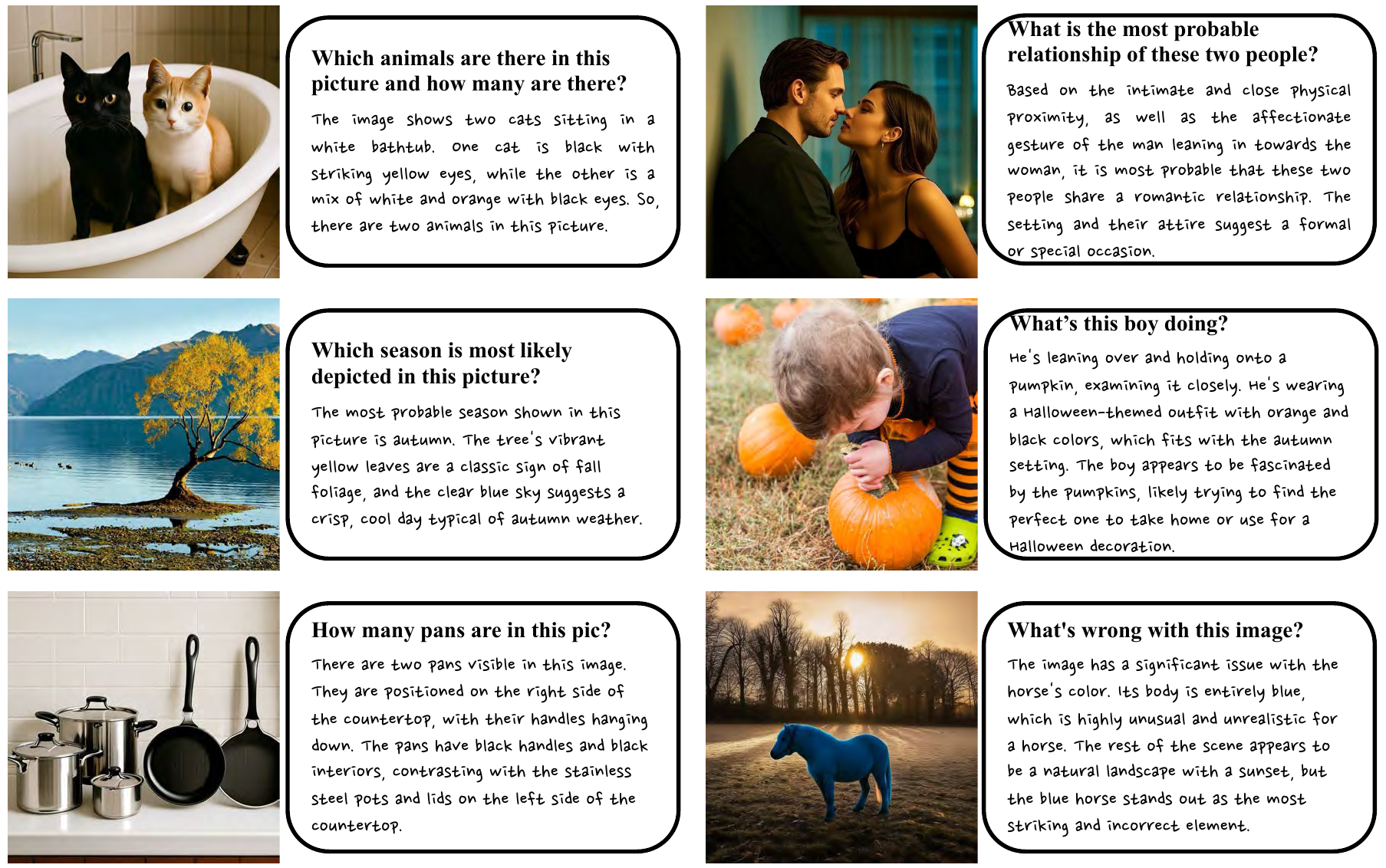}
    \caption{More qualitative results from Ovis-U1 on image understanding.}
    \label{fig:qualitative_iu}
\end{figure}

\subsection{More qualitative results}

We present comprehensive qualitative results in  Fig.~\ref{fig:qualitative_t2i},~\ref{fig:qualitative_ie} and~\ref{fig:qualitative_iu}, to demonstrate Ovis-U1's capabilities across multimodal tasks.

As shown in Fig.~\ref{fig:qualitative_iu}, the model exhibits robust reasoning by interpreting complex visual content and generating contextually coherent responses. It excels at identifying fine-grained details, such as object attributes, spatial hierarchies, and subtle interactions while maintaining contextual awareness for domain-specific tasks. 

Fig.~\ref{fig:qualitative_t2i} highlights the model’s ability to produce high-fidelity images with diverse aesthetics and structural coherence. Ovis-U1 synthesizes visually striking outputs spanning photorealistic scenes, abstract concepts, and hybrid designs, while preserving intricate textures. It performs well even under complex prompts involving multi-object arrangements, spatial constraints, or abstract attribute bindings.

Fig.~\ref{fig:qualitative_ie} demonstrates Ovis-U1’s precision in localized modifications while retaining background integrity. The model executes content replacement, stylistic transformations, and structural edits with minimal artifacts, adhering strictly to instructional prompts.

These qualitative results, paired with quantitative evaluations presented previously, position Ovis-U1 as a versatile foundation for multimodal generative tasks. Its compact 3.6 B parameter architecture balances efficiency and scalability, offering strong potential for performance gains through larger-scale training while maintaining practical deployment viability.

\section{Conclusion}
\label{sec:conclusion}

In this report, we present Ovis-U1, a 3-billion-parameter unified model that excels in multimodal understanding, text-to-image generation, and image editing. As the initial version in the Ovis unified model series, this report addresses key foundational challenges: the design of the visual decoder, its connector with large language models, and the comprehensive training procedure for the unified model. We emphasize the critical role of unified training in aligning the visual encoder, which significantly enhances both understanding and generation performance through collaborative training.
Moreover, we utilize a robust evaluation framework for assessing the unified model's capabilities. We have curated widely-accepted benchmarks in the fields of understanding, text-to-image generation, and image editing to ensure comprehensive evaluation. With only 3B parameters, Ovis-U1 demonstrates strong performance across these benchmarks, even surpassing some task-specific models. This achievement underscores Ovis-U1's ability to advance the boundaries of unified model capabilities.

In the future, we will focus on advancing the powerful unified models.
First, we plan to expand our model by increasing the number of parameters. In the realm of image generation, smaller models often struggle with artifacts and hallucinations. By incorporating more parameters, the model can mitigate these issues and produce higher quality images.
Second, we will enhance our training data pipeline by collecting and curating more diverse, high-quality datasets specifically designed for unified model training, with particular emphasis on interleaved image-text content.
Third, we plan to innovate architectural designs tailored for unified models. To enhance image editing capabilities, we will implement specialized visual encoder-decoder structures optimized to preserve fine-grained details from input images.
Last but not least, we acknowledge that Ovis-U1 currently lacks a reinforcement learning stage, which has proven crucial for large model optimization. Developing effective methods to align unified multimodal models with human preferences remains an important open research question in this domain.

\section{Contributors}

Guo-Hua Wang\footnote{Correspondence to Guo-Hua Wang $<$\texttt{wangguohua@alibaba-inc.com}$>$}, Shanshan Zhao, Xinjie Zhang, Liangfu Cao, Pengxin Zhan, Lunhao Duan, Shiyin Lu, Minghao Fu, Xiaohao Chen, Jianshan Zhao, Yang Li, Qing-Guo Chen

\clearpage

\bibliographystyle{colm2024_conference}
\bibliography{colm2024_conference}

\end{document}